\documentclass[runningheads]{llncs}
\usepackage[T1]{fontenc}
\usepackage[babel=true,expansion=true,kerning=true,tracking=true,spacing=true,verbose=silent]{microtype}
\usepackage{graphicx}
\usepackage{amsmath}
\usepackage{amssymb}
\usepackage{dsfont}
\usepackage{booktabs}
\usepackage{multirow}
\usepackage{siunitx}
\usepackage{algorithm}
\usepackage{algpseudocode}
\newcommand{\Input}{\item[\textbf{Input:}]}
\newcommand{\Output}{\item[\textbf{Output:}]}
\usepackage{xcolor}
\usepackage{xurl}
\usepackage{caption}
\usepackage{subcaption}
\usepackage[rightcaption]{sidecap}
\usepackage{hyperref}  
\ifdefined\nohyperref\else\ifdefined\hypersetup
  \definecolor{mydarkblue}{rgb}{0,0.08,0.45}
  \hypersetup{ %
    pdfborder=0 0 0,
    pdfpagemode=UseNone,
    colorlinks=true,
    linkcolor=mydarkblue,
    citecolor=mydarkblue,
    filecolor=mydarkblue,
    urlcolor=mydarkblue,
    }
  \fi
\fi

\usepackage[capitalize,nameinlink]{cleveref}
\makeatletter
\newcommand{\@chapapp}{\relax}%
\g@addto@macro\appendix{%
}
\makeatother
\usepackage[title]{appendix}
\crefname{appendix}{Appendix}{Appendices}
\Crefname{appendix}{Appendix}{Appendices}
\crefname{section}{Sec.}{Secs.}
\Crefname{section}{Sec.}{Secs.}
\crefname{algorithm}{Alg.}{Algs.}
\Crefname{algorithm}{Alg.}{Algs.}

\renewcommand{\bbbone}{\mathds{1}}

\begin{document}
\title{Bias Leaves a Gradient Trail:\\Label-Free Bias Identification via Gradient Probes on Concept Decompositions}
\titlerunning{Bias Identification via Gradient Probes on Concept Decompositions}
\author{Thomas Vitry\inst{1,2, *}\orcidID{0009-0008-3329-2026} \and
Kieran Edgeworth\inst{1}\orcidID{0009-0008-0310-7088} \and
Stefan Wermter\inst{1}\orcidID{0000-0003-1343-4775} \and
Jae Hee Lee\inst{1, *}\orcidID{0000-0001-9840-780X}}
\authorrunning{T. Vitry et al.}
\institute{University of Hamburg, Hamburg, Germany \and
Ecole Normale Superieure de Rennes, Rennes, France
\email{\{thomas.vitry,kieran.edgeworth\}@studium.uni-hamburg.de}\\
\email{\{stefan.wermter,jae.hee.lee\}@uni-hamburg.de}
}
\maketitle              %

\def\thefootnote{*}\footnotetext{Corresponding authors}\def\thefootnote{\arabic{footnote}}
\begin{abstract}
Vision classifiers can exploit spurious correlations, achieving high in-distribution accuracy yet failing under distribution shift. Existing approaches to bias mitigation and analysis often depend on curated datasets, spurious-attribute or group labels, or retraining, which may be infeasible once a model is deployed or the relevant bias is unknown. We present a bias-label-free, post-hoc method for identifying spurious concepts in frozen vision models, relying only on standard class labels from a held-out audit dataset. For each target class, we collect patches from inputs predicted as that class and apply non-negative matrix factorization to intermediate activations to obtain a bank of interpretable concept vectors. Candidate concepts are then ranked with a bias estimator derived from their interaction with backpropagated gradients on misclassified examples: bias concepts tend to get activated when correcting false negatives and suppressed when correcting false positives. On Colored MNIST and Waterbirds the method recovers concepts aligned with the known spurious cue, and on CelebA it surfaces decision-relevant directions that only partially coincide with the annotated gender attribute; suppressing the top-ranked concepts at inference time improves worst-group accuracy by up to $17.9$ percentage points on Waterbirds and $10.4$ on CelebA without any retraining or parameter updates. Our method identifies decision-relevant spurious directions that need not coincide with annotated ones, providing both an interpretable auditing tool and an actionable debiasing handle for frozen vision models. Code is available at \url{https://github.com/vitryt/label-free-bias-identification}.

\end{abstract}
\section{Introduction}
\label{sec:introduction}

Modern vision models can achieve high in-distribution accuracy while relying on spurious correlations: attributes that predict the label in the training data but are not causally related to the task. This shortcut learning behavior~\cite{geirhos_shortcut_2020} can lead to brittle generalization and systematic failures when the correlation shifts at deployment. For example, in fine-grained recognition the model may over-rely on context: on Waterbirds~\cite{sagawa_distributionally_2019}, background type is spuriously correlated with the bird label during training. Such shortcuts may remain hidden when the available held-out set (e.g., the validation set) shares the same correlation, yet cause systematic errors when the correlation breaks at deployment.

To counter shortcut learning, methods often mitigate spurious correlations during data collection or training~\cite{lee_improving_2023}. When the relevant bias concept is known, it can be targeted for mitigation via retraining~\cite{bahng_learning_2020,correa_efficient_2024,dreyer_hope_2024,kim_learning_2019} or post-hoc interventions on inputs or representations~\cite{anders_finding_2022}. These pipelines typically assume that the spurious attribute has already been identified and therefore do not solve the discovery problem.  In practice, however, we only have access to a deployed frozen model and a held-out set, but not to the original training pipeline, and rarely to bias or group annotations. In this setting, analyzing a deployed model is difficult because the spurious attribute is often unknown, and collecting new annotations can be costly. This motivates methods that can propose plausible spurious attributes post-hoc, while returning a concept vector in the latent space that is directly compatible with downstream interventions.

\begin{figure*}[tb]
	\centering
	\includegraphics[trim={15pt 0pt 15pt 0pt},clip,width=\textwidth]{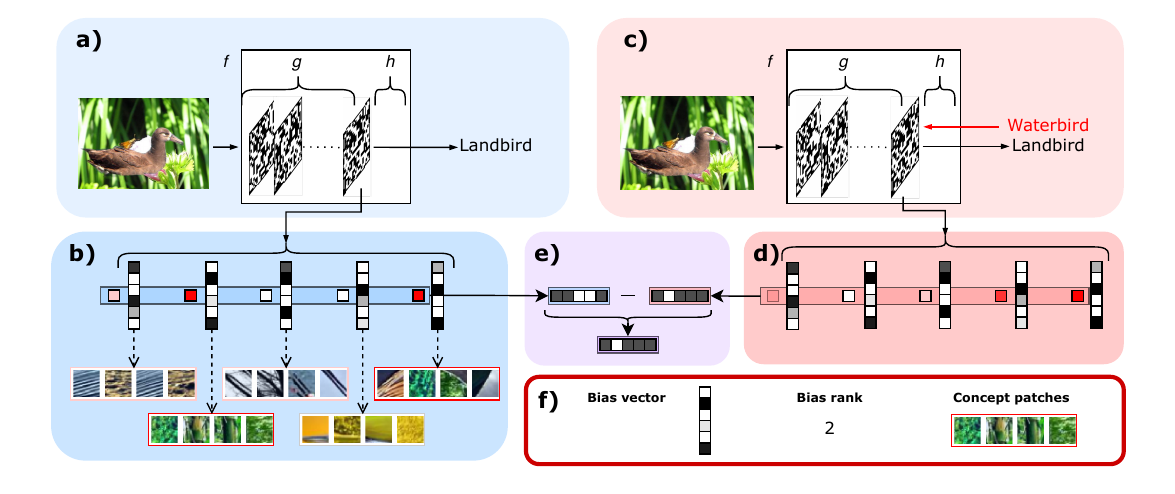}
	\caption{Diagram of the bias identification method. \textbf{(a)} We collect false negatives and false positives from a held-out set (the \emph{bias-audit} set) and pass them through the model. Here, studying class $y{=}0$ (landbird), the sample is a false positive: a waterbird misclassified as a landbird. \textbf{(b)} Using non-negative concept decomposition trained beforehand on the bias-audit set, we decompose activations of the sample into its concept coefficients. \textbf{(c)} Using the ground-truth label from the bias-audit set, we backpropagate the gradient to the activations. \textbf{(d)} We decompose the updated activations (after the gradient step) using the same concept bank. \textbf{(e)} We compare which concepts are activated before and after the gradient step. Concepts that are removed (for false positives) or added (for false negatives) by the gradient step are candidate bias concepts. \textbf{(f)} The identified bias concepts come with their encoding in the activation space and their most activating patches, enabling interpretation. Note that bias concepts are identified by aggregating over multiple samples, not from a single example.}
	\label{fig:identification_method}
\end{figure*}

We study whether unsupervised, representation-level concept discovery can detect spurious attributes in frozen vision models when bias annotations are unavailable. Our approach decomposes intermediate activations into non-negative concept vectors and uses a held-out set (without group or bias labels) to probe the model via gradients (\cref{sec:identification_method}). We first test whether the spurious attribute learned by a biased model reliably appears as a distinct concept direction in this basis (\cref{sec:bias_align}). Building on this, we propose a bias identification criterion based on how concept activations change under a gradient descent step: bias concepts tend to be \emph{activated} when correcting false negatives and \emph{suppressed} when correcting false positives, while concepts encoding intrinsic class evidence change much less asymmetrically. We validate the scoring criterion against ground-truth bias (\cref{sec:bias_ident,sec:correlation}) and then test whether the identified concepts are actionable, by suppressing them at inference time without any parameter updates (\cref{sec:debiasing}). We evaluate these on Colored MNIST, Waterbirds and CelebA, three standard benchmarks for spurious correlation learning, and find that concept suppression substantially improves worst-group accuracy on Waterbirds and CelebA. The resulting approach provides both an interpretable auditing tool and a concrete debiasing handle for frozen vision models when bias annotations and retraining are unavailable.

\section{Preliminaries}
\label{sec:preliminaries}

We consider a multi-class classification task with inputs $x\in\mathcal{X}\subseteq\mathbb{R}^d$ and labels $y\in\mathcal{Y}=\{1,\dots,C\}$, and a hypothetical ideal classifier $f^*:\mathcal{X}\to\mathcal{Y}$. Following~\cite{lee_improving_2023}, an \emph{attribute} is a function $b:\mathcal{X}\to\{0,1\}$; it is \emph{intrinsic} to class $y$ if it defines membership (e.g., digit shape), and a \emph{bias attribute} for $y$ if it is spuriously correlated with $y$ without being intrinsic. We are given a frozen classifier $f$ with class scores $f(x)=(f_c(x))_{c=1}^C$ and a held-out \emph{bias-audit} set $\mathcal{D}=\{(x_i,y_i)\}_{i=1}^n$ with $y_i=f^*(x_i)$ but no attribute or group annotations. $\mathcal{D}$ need not be from the training set and may be unbiased or distribution-shifted; it may come from a validation split, relabeled deployment data, or targeted stress tests. Our goal is to detect, for each class $y$, concept directions in the representation space that plausibly encode bias attributes used by the model and that are empirically implicated in its decisions.

\paragraph{Non-Negative Concept Decomposition.}
Following CRAFT~\cite{fel_craft_2023}, we write $f=h\circ g$ where $g:\mathcal{X}\to\mathcal{A}$ maps an input to an intermediate representation $a=g(x)\in\mathcal{A}\subseteq\mathbb{R}^p$ and $h:\mathcal{A}\to\mathbb{R}^C$ maps the representation to class scores. We choose a representation layer after a ReLU so that $a\in\mathbb{R}^p_{\ge 0}$, enabling non-negative matrix factorization (NMF). For each class $y$, let $\hat{y}(x)=\arg\max_{c\in\mathcal{Y}} f_c(x)$ be the predicted label and collect inputs predicted as $y$:
\[
	\mathcal{X}_y=\{x_i : \hat{y}(x_i)=y\}.
\]
A patch dataset is then created by applying a crop-and-resize operator $\pi_s$ (patch size $s$) to these images, yielding patches $\mathcal{P}_y=\{\pi_s(x) : x\in\mathcal{X}_y\}$. Let $A_y=g(\mathcal{P}_y)\in\mathbb{R}_{\ge 0}^{n_y\times p}$ be the resulting activations. We compute a class-conditional concept bank
\begin{equation}
	\label{eqn:craft}
	(U_y, W_y)=\underset{U_y\ge 0,\,W_y\ge 0}{\text{arg}\,\text{min}} \frac{1}{2}\left\lVert A_y-U_yW_y^\top\right\rVert_F^2,
\end{equation}
via NMF, where $W_y\in\mathbb{R}_{\ge 0}^{p\times r}$ contains $r$ concept vectors $w_{y,1},\dots,w_{y,r}$ and $U_y\in\mathbb{R}_{\ge 0}^{n_y\times r}$ the corresponding non-negative concept coefficients. As in~\cite{fel_craft_2023}, this typically yields semantically meaningful, relatively disentangled concepts; the decomposition is approximate and its residual is captured by~\cref{eqn:craft}. In practice, we solve this NMF objective by alternating between two non-negative least-squares (NNLS) subproblems, fixing $W_y$ while optimizing $U_y$ and vice versa. This monotonically decreases the objective and converges to a stationary point under standard assumptions.

Given $W_y$, we obtain concept coefficients for any representation vector $a\in\mathbb{R}^p$ (in particular, $a=g(x)\in\mathbb{R}_{\ge 0}^p$) by solving an NNLS problem,
\begin{equation}
	u_y(a)=\underset{u\ge 0}{\text{arg}\,\text{min}}\left\lVert a-uW_y^\top\right\rVert_2^2, \label{eq:nnls}
\end{equation}
and write $u_{y,k}(a)$ for its $k$-th component.

\section{Gradient-Based Bias Identification}
\label{sec:identification_method}

We hypothesize that spurious attributes learned by a biased model appear as distinct concept directions in the non-negative concept decomposition, and that a gradient-based criterion can identify them among the extracted concepts.

Our intuition is as follows. A partially biased model uses both intrinsic and spurious attributes. Intrinsic concepts for class $y$ should be active on true positives, while bias concepts need not be. On a false negative of class $y$, missing evidence is therefore more likely due to missing spurious support than missing intrinsic evidence; on a false positive, intrinsic evidence is typically absent and the error is more likely driven by a spurious attribute. Since NMF enforces non-negativity, evidence can only be added via positive concept activations. Because activity alone is noisy, we use gradient information to isolate the concepts the model itself would add or remove to correct its prediction: by propagating the loss gradient to activations (instead of updating weights), we measure which concepts change under one probe step. A bias concept should be systematically absent then added for false negatives, and present then removed for false positives.

To identify bias concepts without access to $b(x)$, we therefore measure how concept coefficients change under a single gradient probe step that improves the model's prediction on a given example. For a labeled example $(x_i,y_i)$ with representation $a=g(x_i)$, let $L(h(a),y_i)$ be the cross-entropy loss and compute the gradient w.r.t.\ the representation, $\nabla_a L(h(a),y_i)$. We define a perturbed representation
\begin{equation}
	\label{eqn:probe_update}
	a' = a - d\,\nabla_a L(h(a),y_i),
\end{equation}
with gradient step size $d>0$. After making $a'$ non-negative, we compute $u_{y,k}(a')$ via the same non-negative least squares problem. We treat a concept as \emph{active} if its coefficient is positive and write
\[
	I_{y,k}(a)=\bbbone[u_{y,k}(a)>0],
\]
where $\bbbone[\cdot]$ is the indicator function.

For a class $y$, we define false negatives and false positives with respect to $y$:
\[
	\mathrm{FN}_y = \{(x_i,y_i) \mid y_i=y,\,\hat{y}(x_i)\neq y\},\qquad
	\mathrm{FP}_y = \{(x_i,y_i) \mid \hat{y}(x_i)=y,\,y_i\neq y\}.
\]
More explicitly, we define the false-negative and false-positive estimators for concept $k$ as:
\[
	E_{\mathrm{FN}}^{y,k} =\hspace{-1em}\sum_{(x_i,y_i)\in\mathrm{FN}_y}\hspace{-1em}\frac{I_{y,k}(a_i') - I_{y,k}(a_i)}{|\mathrm{FN}_y|},\qquad
	E_{\mathrm{FP}}^{y,k} = \hspace{-1em}\sum_{(x_i,y_i)\in\mathrm{FP}_y}\hspace{-1em}\frac{I_{y,k}(a_i) - I_{y,k}(a_i')}{|\mathrm{FP}_y|},
\]
where $a_i=g(x_i)$ and $a_i'$ is obtained from \cref{eqn:probe_update}. The two terms measure how consistently concept $k$ is \emph{added} on false negatives and \emph{removed} on false positives under the gradient step.

\begin{SCfigure}[1.5][t]
	\centering
	\includegraphics[trim={15pt 0pt 15pt 20pt},clip,width=0.4\linewidth]{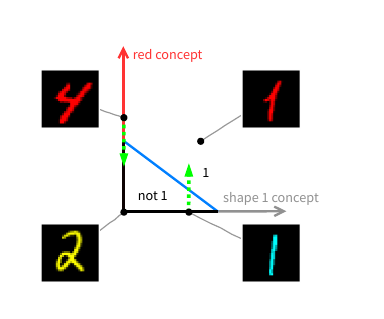}
	\caption{Intuition of the gradient-concept interaction on class $y$ ($=1$). The vertical axis is the bias concept \texttt{red} and the horizontal axis is the relevant concept \texttt{shape~1}. The blue line is the model's decision boundary. For false negatives, i.e., image (\texttt{cyan},~\texttt{1}), the gradient step increases bias attribute; for false positives, i.e., image (\texttt{red}, \texttt{4}), the same probe removes the bias attribute. Intrinsic attribute, i.e., \texttt{shape 1}, is expected to show a weaker asymmetric pattern.}
	\label{fig:spurious_intuition}
\end{SCfigure}

As illustrated in \cref{fig:spurious_intuition}, we hypothesize that (i) bias concepts are more likely to increase under the gradient probe update in \cref{eqn:probe_update} on $\mathrm{FN}_y$ (i.e., the model prefers the spurious cue to predict $y$) and decrease on $\mathrm{FP}_y$ (i.e., reducing features that spuriously support $y$), and (ii) that this asymmetric behavior does \emph{not} hold strongly for intrinsic concepts, i.e., false negatives possess intrinsic attributes of class $y$ (otherwise they would not have ground-truth label $y$), and false positives lack them. More precisely, for false negatives we expect the probability of a bias concept $k$ being activated to increase after the gradient step:
\[
	\mathbb{P}\bigl(I_{y,k}(a') = 1\bigr) - \mathbb{P}\bigl(I_{y,k}(a) = 1\bigr) > 0,
\]
whereas for false positives we expect the \emph{opposite}.
To rank concepts, we define the bias score of concept $k$ for class $y$ as
\begin{equation}
	\label{eqn:score}
	S_{y,k}=\frac{1}{2}\left(E_{\mathrm{FN}}^{y,k}+E_{\mathrm{FP}}^{y,k}\right).
\end{equation}

For each example in $\mathrm{FN}_y\cup\mathrm{FP}_y$, scoring requires one backward pass to compute $\nabla_a L$ and two non-negative least squares solves to obtain $u_y(a)$ and $u_y(a')$. The non-negative concept decomposition is computed once per class. \Cref{fig:identification_method} gives a visual overview of the pipeline, and \cref{alg:bias_identification} in \cref{app:algorithm} summarizes the full procedure, from class-conditional concept extraction to bias scoring and ranking.

\section{Experiments}
\label{sec:expe}

We evaluate our method on standard benchmarks for spurious correlation learning and the hypotheses from \cref{sec:identification_method}.

\subsubsection{Datasets.}
We use three standard spurious-correlation benchmarks with a $0.95$ train-time correlation. \textbf{CMNIST} is a colored variant of MNIST where color acts as the spurious attribute (implementation adapted from~\cite{bahng_learning_2020}); we also construct an \emph{unbiased} variant and a \texttt{shifted-bias} variant as distribution-shifted audit sets. \textbf{Waterbirds}~\cite{sagawa_distributionally_2019} classifies waterbirds vs.\ landbirds~\cite{wah_caltech-ucsd_nodate} with water/land background~\cite{zhou_places_2018} as the spurious attribute. \textbf{CelebA}~\cite{liu2015faceattributes} is a blond-hair vs.\ not-blond binary task with gender as the spurious attribute~\cite{Wu2023CelebA}; it is particularly challenging due to a severe class imbalance in the test split (only $180$ blond men vs.\ $9767$ not-blond women), making overall accuracy an unreliable metric. Split construction follows standard practice (details in \cref{app:datasets}).

\subsubsection{Audit bias.}
In all experiments, we vary the bias structure of the bias-audit set $\mathcal{D}$, a categorical we refer to as the \emph{audit bias} throughout. We consider three values: \texttt{biased} (same spurious correlation as the training distribution), \texttt{unbiased} (correlation broken), and \texttt{shifted bias} (a different spurious assignment than training). CMNIST evaluates all three; Waterbirds and CelebA are only available in biased form, so we restrict them to \texttt{biased}. For all audit variants, the underlying images are taken from the held-out validation split of the corresponding dataset.
\subsubsection{Models and Training.}
For CMNIST, we use a three-layer MLP with ReLU activations, trained for 100 epochs with stochastic gradient descent (SGD) and batch size 128. For the gradient descent, we pick a learning rate of $0.01$ halved every $25$ epochs. Waterbirds experiments are run on a ResNet-18 with pretrained weights~\cite{he_deep_2016}. The model is trained with SGD (learning rate $10^{-1}$ divided by $10$ every 30 epochs) for $100$ epochs with batch size $1024$. Finally, we use a ResNet-50 with pretrained weights \cite{he_deep_2016} trained again with SGD (learning rate $10^{-4}$) for $20$ epochs with batch size $512$ for CelebA. For non-negative concept decomposition, we extract activations from the final ReLU layer before the classifier head to ensure non-negativity. Experiments are repeated 10 times with different random seeds; we report means and standard errors.

\subsection{Concept--Bias Alignment}
\label{sec:bias_align}

We first investigate whether the spurious attributes learned by the model appear as distinct concept directions in the non-negative concept decomposition. We hypothesize that if the model relies on a spurious attribute to predict class $y$, then at least one concept in $W_y$ aligns with that attribute. To test this, we use datasets with known ground-truth spurious attributes. On CMNIST, we measure cosine similarity between each concept vector and an estimated color-bias direction following~\cite{dreyer_hope_2024}, treating concepts with similarity $\ge 0.55$ as bias-aligned. On Waterbirds, we compute the average fraction of \emph{background} pixels in each concept's most activating patches using foreground masks; concepts with $\ge 85\%$ background pixels are considered bias-aligned. We sweep the patch size $s$ and number of concepts $r$.

\begin{figure}[tb]
	\centering
	\begin{subfigure}[t]{0.45\linewidth}
		\centering
		\includegraphics[height=4cm]{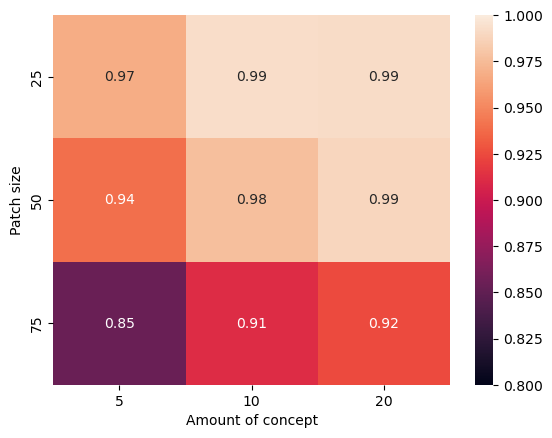}
		\caption{Bias-alignment score (fraction of background pixels in most-activating patches) of the most background-dominated concept as a function of patch size and number of concepts (both classes, 5~runs per class).}
		\label{fig:waterbirds_alignment_app}
	\end{subfigure}\hfill
	\begin{subfigure}[t]{0.52\linewidth}
		\centering
		\includegraphics[height=4cm]{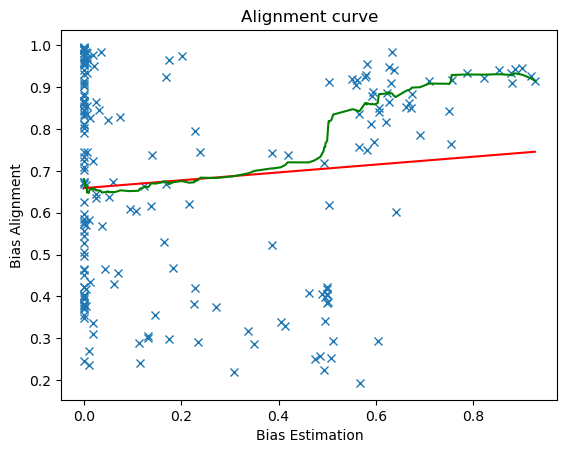}
		\caption{Waterbirds bias score against bias-alignment score. Blue crosses represent concepts, red curves are linear regressions, and green curves show the average bias-alignment score of concepts above the current bias score threshold.}
		\label{fig:waterbirds_bias_score_correlation}
	\end{subfigure}
	\caption{Waterbirds concept bias analysis.}
\end{figure}
\subsubsection{Results.}
On Waterbirds (\cref{fig:waterbirds_alignment_app}), we reliably find a background-dominated concept with very low variance (below $2\%$ standard error), confirming that a bias-aligned concept is consistently recovered. The best results are obtained with patch size around $50$ and $r{\geq}10$: patches at this scale capture enough background structure to form a coherent concept while remaining small relative to the full image, and sufficiently many concepts are needed for a dedicated bias direction to emerge. CMNIST experiments confirm these findings across multiple audit-set distributions and show that unbiased models do not produce spurious concept directions (see \cref{app:cmnist}). These experiments confirm that non-negative concept decomposition reliably captures the bias attribute in biased models. Balancing interpretability and decomposition quality, we set ($r = 8$, $s = 6$) for CMNIST and ($r = 10$, $s = 50$) for Waterbirds and CelebA; gradient step size is set to $d=2\times10^{4}$ (sensitivity analysis in \cref{app:cmnist}).

\subsection{Bias Score Validation}
\label{sec:bias_ident}

\begin{figure}[tb]
	\centering
	\includegraphics[width=0.85\linewidth]{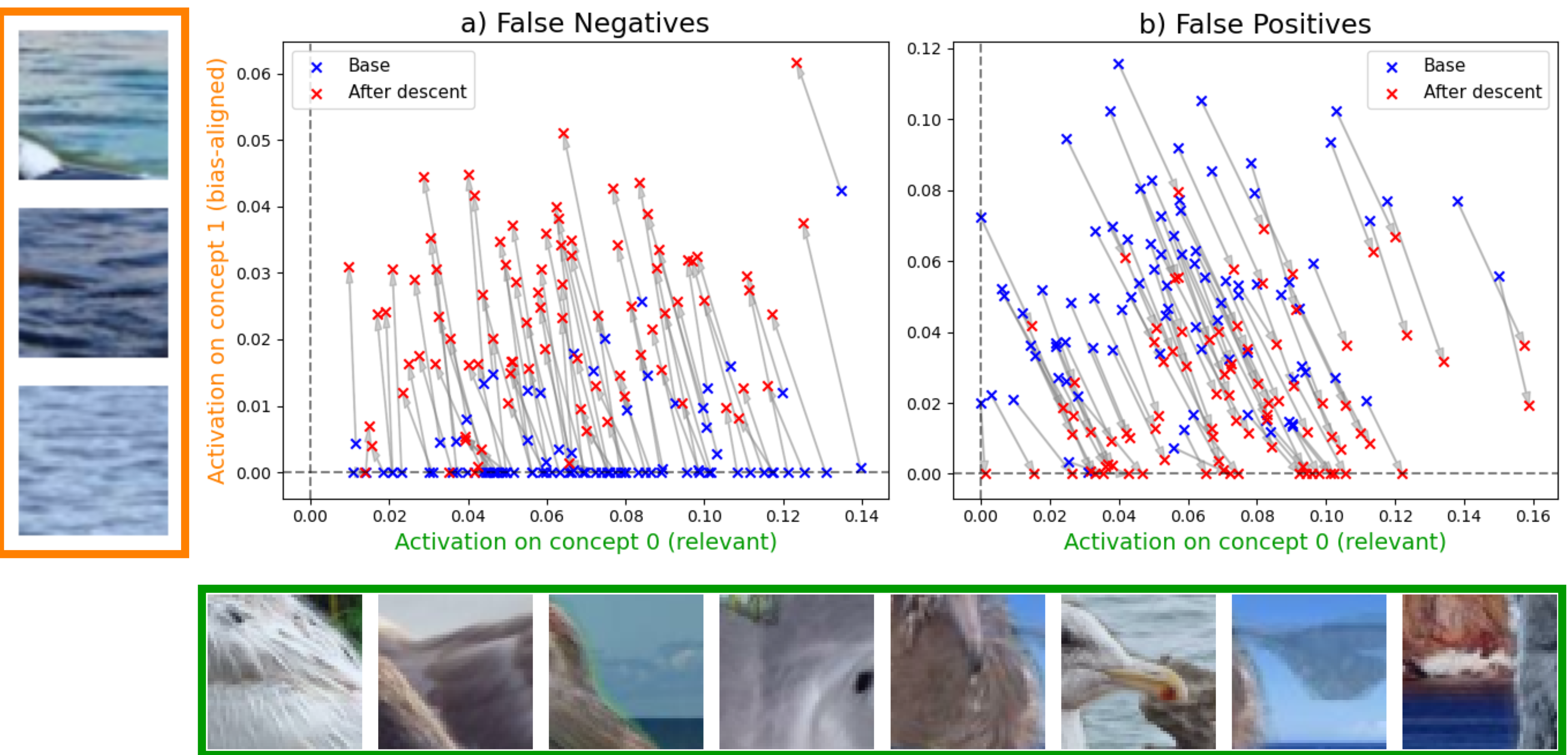}
	\caption{Empirical validation of the gradient-concept interaction on a ResNet-18 trained on Waterbirds, to be compared with \cref{fig:spurious_intuition}. Each panel scatters concept activations for the waterbird class, with the relevant concept (concept~0, bird features; green frame) on the horizontal axis and the bias-aligned concept (concept~1, sea backgrounds; orange frame) on the vertical axis. Blue crosses: base activations $u_y(a)$; red crosses: activations $u_y(a')$ after one gradient probe step (\cref{eqn:probe_update}); grey segments connect the two. \textbf{(a)}~On false negatives, the bias concept is systematically added. \textbf{(b)}~On false positives, it is systematically suppressed. The relevant concept shows a markedly weaker and less asymmetric change.}
	\label{fig:hypothese_validated}
\end{figure}

Having confirmed that bias directions emerge in the NMF basis, we now test whether our scoring criterion ranks them automatically. We apply the gradient-concept interaction score from \cref{eqn:score} to rank concepts as candidate bias directions on misclassified and distribution-shifted examples, and evaluate if the high-ranked concepts correspond to the ground-truth bias concepts from \cref{sec:bias_align}. \Cref{fig:hypothese_validated} first illustrates this interaction directly on Waterbirds: for a bird-focused relevant concept and a sea-focused bias concept, the gradient probe step consistently increases the bias-concept activation on false negatives and decreases it on false positives, empirically confirming the intuition of \cref{fig:spurious_intuition} on real data.

\Cref{fig:waterbirds_bias_score_correlation} further demonstrates that the relation between bias score and bias alignment is strong: concepts with bias scores above $0.55$ are reliably background-dominated (mean alignment above $0.8$), supporting the use of this threshold for bias identification. On CMNIST, the same pattern holds for biased models, while unbiased models show no strong correlation and no concept exceeds a bias score of $0.7$ (see \cref{app:cmnist}). Across both datasets, concepts with bias scores above $0.55$ are consistently aligned with the ground-truth bias. We therefore use a bias score threshold of $\tau = 0.55$ for the remainder of this study; this threshold can be increased to reduce the false-positive rate at the cost of sensitivity.

\subsection{Correlation with Ground-Truth Bias}
\label{sec:correlation}
To quantitatively validate that the identified concepts correspond to the ground-truth bias, we measure the correlation between concept activations and ground-truth bias labels on the test sets. For each concept in the merged concept bank and each bias attribute, we compute the absolute Matthews Correlation Coefficient (MCC) between the binarized concept activation $I_{y,k}$ and the binary bias label.\footnote{We use the absolute value because the sign of a concept direction under NMF is determined by the factorization and is not semantically meaningful: a concept that anti-correlates with the binary bias label encodes the same bias information as one that correlates with it.} The absolute MCC values of identified bias concepts with their corresponding bias attribute are then compared against those of all other concept--bias pairs. We report the proportion of statistically significant pairs under a $\chi^2$ independence test and conduct a one-sided Mann--Whitney $U$-test to assess whether identified bias concepts exhibit significantly higher correlation than other pairs. Results are reported in \cref{tab:correlation_metrics}.

\begin{table}[tb]
\centering
\small
\caption{Correlation between identified bias concepts and ground-truth bias labels. \#bias: number of identified bias concepts; $f_{\text{bias}}$ / $f_{\text{other}}$: proportion of statistically significant pairs ($\chi^2$ test) for bias concepts vs.\ other concept--bias pairs; $\bar{\Phi}_{\text{bias}}$ / $\bar{\Phi}_{\text{other}}$: mean absolute MCC of significant pairs; $p$: one-sided Mann--Whitney $U$-test. CMNIST results are reported in \cref{app:cmnist}.\label{tab:correlation_metrics}}
\setlength{\tabcolsep}{.25em}
\renewcommand{\arraystretch}{1.05}
\resizebox{\linewidth}{!}{%
\begin{tabular}{@{}lcS[table-format=1.1(1)]S[table-format=1.3]S[table-format=1.3]S[table-format=-1.3(3)]S[table-format=-1.3(3)]c@{}}
\toprule
{Train dataset} & {Audit bias} & {\#bias} & {$f_{\text{bias}}$} & {$f_{\text{other}}$} & {$\bar{\Phi}_{\text{bias}}$} & {$\bar{\Phi}_{\text{other}}$} & {$p$} \\
\midrule
Waterbirds & biased & ${4.3{\scriptstyle\pm0.6}}$ & ${0.977}$ & ${0.790}$ & ${0.389{\scriptstyle\pm0.175}}$ & ${0.154{\scriptstyle\pm0.123}}$ & ${0.001}$ \\
CelebA & biased & ${3.2{\scriptstyle\pm0.9}}$ & ${0.969}$ & ${0.768}$ & ${0.163{\scriptstyle\pm0.076}}$ & ${0.124{\scriptstyle\pm0.083}}$ & ${0.010}$ \\
\bottomrule
\end{tabular}}
\end{table}

On Waterbirds, activations of identified bias concepts are highly correlated with the background attribute ($f_{\text{bias}} = .977$, $p < 0.001$), consistently exceeding other concept--bias pairs. On CelebA, identified concepts ($f_{\text{bias}} = .969$) are statistically associated with bias but MCC values remain low, and no concept exceeds an MCC of $0.4$. The Mann--Whitney test still shows a significant advantage over other concepts ($p = 0.010$), indicating that the identified concepts are relatively more bias-aligned than the remaining ones. The low absolute values suggest that the single-scale NMF decomposition struggles to capture a high-level attribute such as gender, possibly compounded by the severe class imbalance (blond:not blond $\approx$ $1{:}6.5$). CMNIST results (\cref{app:cmnist}) confirm strong correlation for biased models and no correlation for unbiased models.

\subsection{Inference-Time Bias Mitigation}
\label{sec:debiasing}

To test whether identified bias directions are actionable, we apply a simple inference-time concept suppression inspired by projection-based methods such as P-ClArC~\cite{anders_finding_2022}. Adapting such projection-style interventions to NMF-based concept banks raises two practical considerations. First, our bias-identification stage produces class-conditional concept banks, meaning each decomposition is tailored to one class and is therefore not directly suitable for full downstream prediction. We address this by merging class-wise concept banks into a model-wide bank $W_{\text{merged}}$ and clustering highly similar concepts (cosine similarity above $0.95$). Second, projection-based mitigation requires a neutral value to which the suppressed concept is set. Since NMF matrices are naturally sparse, we use $0$ as a neutral value for each concept, avoiding the need to estimate concept-specific baselines from additional data. For a test input $x$, we compute its representation $a=g(x)$ and concept coefficients $u_{\text{merged}}(a)$, and remove the components along bias concepts $B$ before rescaling the result to the original activation norm and performing the classification:
\[
	a_{\text{sup}} = a - \sum_{k\in B} u_{\text{merged}}(a)_k\,w_{\text{merged},k},
	\qquad
	a_{\text{rsc}} = a_{\text{sup}}\,\frac{\Vert a\Vert_2}{\Vert a_{\text{sup}}\Vert_2}
	\qquad
	\tilde{f}(x)=h(a_{\text{rsc}}),
\]
where $a_{\text{sup}}$ is the bias-suppressed representation and $a_{\text{rsc}}$ is its rescaling to the original activation norm $\Vert a\Vert_2$, which compensates for the magnitude lost during suppression. Based on the previous experiments, concepts with bias scores above $0.55$ can reliably be treated as bias-aligned. Accordingly, we define $B = \bigcup_{y} \{k:S_{y,k}>\tau\}$ with $\tau=0.55$.
We test this approach on all three datasets. For CMNIST and Waterbirds, we choose hyperparameters based on the previous studies; for CelebA, we reuse the Waterbirds settings because it uses a similar architecture, allowing us to test whether the hyperparameters transfer across datasets. To ensure that effects are linked to our identification, we also run an ablation that randomly suppresses the same number of concepts in the merged bank for each model. These experiments are reported as \textit{ablation} in the results.

\begin{table}[t]
\centering
\small
\caption{Effect of suppressing the identified bias concepts at inference time. Symbol ``--'' under \emph{Audit bias} denotes the base frozen model without mitigation. CMNIST results are reported in \cref{app:cmnist}.\label{tab:debias_metrics}}
\setlength{\tabcolsep}{.25em}
\renewcommand{\arraystretch}{1.05}
\begin{tabular}{@{}llS[table-format=2.1(1.1)]S[table-format=2.1(2.1)]S[table-format=2.1(2.1)]@{}}
\toprule
{Train dataset} & {Audit bias} & {Accuracy} & {Worst-class acc} & {Worst-group acc} \\
\midrule
\multirow{3}{*}{Waterbirds} & -- & ${82.1{\scriptstyle\pm0.3}}$ & ${69.1{\scriptstyle\pm0.8}}$ & ${45.9{\scriptstyle\pm1.5}}$ \\
& biased      & $\mathbf{86.4{\scriptstyle\pm1.7}}$ & $\mathbf{77.6{\scriptstyle\pm2.5}}$ & $\mathbf{63.8{\scriptstyle\pm 4.0}}$ \\
& ablation    & ${84.8{\scriptstyle\pm1.4}}$ & ${66.8{\scriptstyle\pm5.2}}$ & ${43.7{\scriptstyle\pm8.5}}$ \\
\midrule
\multirow{3}{*}{CelebA}     & -- & $\mathbf{95.3{\scriptstyle\pm0.1}}$ & ${81.3{\scriptstyle\pm1.3}}$ & ${43.4{\scriptstyle\pm2.5}}$ \\
& biased      & ${94.9{\scriptstyle\pm0.2}}$ & $\mathbf{86.4{\scriptstyle\pm1.4}}$ & $\mathbf{53.8{\scriptstyle\pm4.1}}$ \\
& ablation    & ${94.1{\scriptstyle\pm1.2}}$ & ${79.3{\scriptstyle\pm11.8}}$ & ${45.4{\scriptstyle\pm17.9}}$ \\
\bottomrule
\end{tabular}
\end{table}

\subsubsection{Results.}
\Cref{tab:debias_metrics} reports overall, worst-class, and worst-group accuracy. On Waterbirds, suppression improves overall accuracy by $+4.3$, worst-class accuracy by $+8.5$, and worst-group accuracy by $+17.9$, clearly outperforming random ablation despite class imbalance. On CelebA, worst-class and worst-group accuracy improve by $+5.1$ and $+10.4$ with only a small drop in overall accuracy and no dataset-specific tuning (Waterbirds settings were reused); the random-suppression ablation ($45.4\pm17.9$ worst-group) is consistent with the effect being concept-specific rather than a pure capacity reduction. Despite the weak correlation with the annotated gender attribute (\cref{sec:correlation}), suppression improves fairness metrics---the qualitative analysis in \cref{sec:interpretation} suggests the identified directions capture decision-relevant spurious information that only partially coincides with gender. On CMNIST, suppression yields mixed results (\cref{app:cmnist}). A detailed comparison with supervised baselines (JTT, Group DRO, DFR) is provided in \cref{app:baselines}.

\subsection{Bias Concept Interpretation}
\label{sec:interpretation}

\begin{figure}[tb]
	\centering
	\begin{subfigure}[t]{.48\linewidth}
		\centering
		\includegraphics[width=\linewidth]{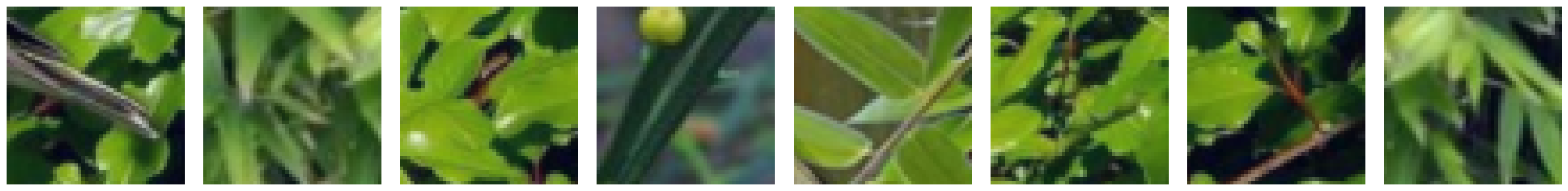}
		\caption{Waterbirds, bias for \texttt{landbird}: \texttt{bamboo}, \texttt{phyllostachys}, \texttt{common bamboo}, \texttt{lesser bullrush}, \texttt{damselfly}.}
        \label{fig:interpret_water_main}
	\end{subfigure}\hfill
	\begin{subfigure}[t]{.48\linewidth}
		\centering
		\includegraphics[width=\linewidth]{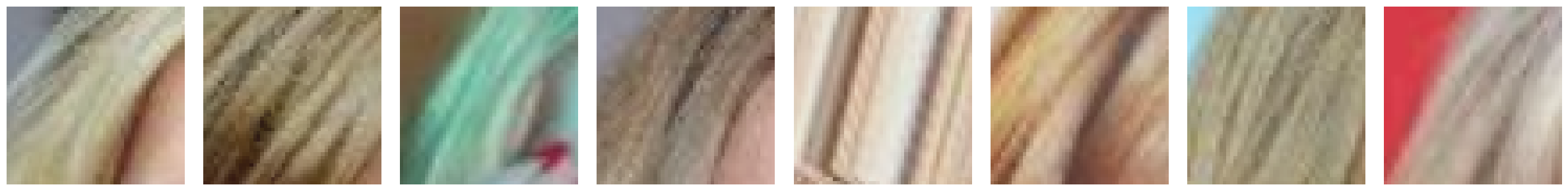}
		\caption{CelebA, bias for \texttt{blond}: \texttt{human hair}, \texttt{hair}, \texttt{body hair}, \texttt{hair space}, \texttt{hair wig}.}
        \label{fig:interpret_celeb_main}
	\end{subfigure}
	\caption{Top-activating patches of representative bias concepts identified by our method, with top-5 generated labels from NOVIC~\cite{allgeuer2025unconstrained}. \textbf{(a)}~Landbird bias concept (vegetation background). \textbf{(b)}~Blond-hair bias concept (hair shape/length). Extended galleries for all concepts are in \cref{app:interpret}.}
    \label{fig:interpret_main}
\end{figure}

Each identified concept comes with its most-activating audit-set patches, which can be inspected qualitatively. To reduce reliance on bias-prone manual inspection, for each bias concept we generate the 10 labels most associated with its top-100 patches (weighted by rank) using NOVIC~\cite{allgeuer2025unconstrained}, an open-vocabulary image classifier; we treat the labels as a qualitative summary rather than ground truth. Representative patch galleries are shown in \cref{fig:interpret_main}; per-dataset extended galleries and discussion are in \cref{app:interpret} (\cref{fig:interpret_cmnist_app,fig:interpret_water_app,fig:interpret_celeb_app}).

On Waterbirds (\cref{fig:interpret_water_main}), the top labels align cleanly with the ground-truth bias: plant-related terms (\texttt{bamboo}, \texttt{phyllostachys}, \texttt{rain tree}) for \texttt{landbird} and marine terms (\texttt{wave}, \texttt{blue whale}, \texttt{sea}) for \texttt{waterbird}. On CelebA (\cref{fig:interpret_celeb_main}), the top labels cluster around hair (\texttt{hair}, \texttt{human hair}, \texttt{hair space}, \ldots); qualitative inspection of the patches suggests the direction captures hair \emph{shape and length} rather than color. This is consistent with the low MCC against the annotated gender attribute (\cref{sec:correlation}): gender on CelebA is a spatially distributed cue that a patch-based decomposition cannot encode as a single direction, but localized proxies such as hair length, eyeshadow, or lips carry partial gender information, which explains why suppressing these concepts still improves worst-group performance (\cref{sec:debiasing}).

\section{Related Work}
\label{sec:related}
Shortcut learning is a pervasive failure mode where models rely on spurious correlations that break under distribution shift~\cite{geirhos_shortcut_2020,ye_clever_2025}. \textbf{Bias-aware mitigation.} When the spurious attribute is known, methods reduce reliance via regularization~\cite{bahng_learning_2020}, adversarial training~\cite{kim_learning_2019}, representation erasure~\cite{belrose_leace_2023-1}, concept-supervised probing~\cite{correa_efficient_2024,kim_interpretability_2018}, or directional interventions~\cite{anders_finding_2022,dreyer_hope_2024}; post-training approaches include saliency-masked fine-tuning~\cite{asgari_masktune_2022} and subspace projection~\cite{holstege_removing_2024}. \textbf{Bias discovery without bias labels.} Related work upweights high-loss examples~\cite{nam_learning_2020,liu_just_2021} or discovers latent failure slices via environment inference and pseudo-labeling~\cite{pezeshki_discovering_2024,zhang_discover_2024,zare_frustratingly_2024,ghosh_ladder_2025,olesen_slicing_2025}, though these generally do not yield a class-conditional direction that can be ranked and targeted~\cite{labonte_group_2024,tsirigotis_group_2023,ghaznavi_exploiting_2025}. Unsupervised concept discovery extracts interpretable directions from activations~\cite{ghorbani_towards_2019,fel_craft_2023,lee_concept-based_2024} and has been applied to robustness~\cite{arefin_unsupervised_2024}, concept ranking~\cite{kowal_understanding_2024-1}, and post-hoc bias proposal via explanation maps or vision-language models~\cite{chakraborty_exmap_2024,kim_discovering_2024,paduraru_conceptdrift_2024}, though often without providing a concrete direction for intervention~\cite{bhusal_face_2025,pahde_navigating_2024-1,panousis_coarse--fine_2024}. \textbf{Post-hoc mitigation with frozen models.} Existing approaches reduce spurious reliance through last-layer retraining on a reweighted split~\cite{kirichenko_last_2023}, sub-network extraction~\cite{le_out_2024}, activation erasure~\cite{he_eva_2024}, data pruning~\cite{mulchandani_severing_2024}, or inference-time debiasing~\cite{gerych_bendvlm_2024,hirota_saner_2024}, but still require optimization, group labels, or model edits. In contrast, we identify class-conditional spurious concept directions in a frozen model without bias labels, providing interpretable handles for downstream intervention without updating parameters. An extended discussion is provided in \cref{app:related}.

\section{Conclusion}
\label{sec:conclusion}

We presented a label-free, post-hoc method for identifying and mitigating spurious concepts in frozen vision classifiers that requires only standard class labels from a held-out audit set. Non-negative concept decomposition reliably recovers spurious attribute directions across hyperparameters on CMNIST and Waterbirds, and our gradient--concept interaction score ranks these directions without any bias annotations. Suppressing the top-ranked concepts at inference time improves worst-group accuracy by up to $17.9$ percentage points on Waterbirds and $10.4$ on CelebA, with no retraining or parameter updates. On CelebA, the identified concepts improve fairness metrics despite correlating only weakly with the annotated gender attribute, indicating that the method captures decision-relevant spurious information that extends beyond pre-defined bias categories. Together, these results demonstrate that unsupervised concept discovery combined with gradient probing can serve both as an interpretable auditing tool and as a practical debiasing handle for deployed models.

\paragraph{Limitations and future work.}
Our approach assumes access to a post-hoc bias-audit set and is most informative when enough misclassified examples are available; in low-error regimes, targeted stress tests or distribution shifts may be needed to surface failures. The single-scale patch decomposition limits the ability to capture bias attributes that are spatially distributed (as observed on CelebA); multi-scale decompositions~\cite{ghorbani_towards_2019} may address this. Bias information may also be entangled across multiple concept directions, leading to collateral suppression of task-relevant features; recursive or hierarchical decompositions could help disentangle these and are a promising direction for future work.

\bibliographystyle{splncs04}
\bibliography{main}

\clearpage
\appendix

\section{Algorithm}
\label[appendix]{app:algorithm}
The bias identification procedure is summarized in \cref{alg:bias_identification}.
\begin{algorithm}[ht]
	\caption{Post-hoc bias identification.}
	\label{alg:bias_identification}
	\begin{algorithmic}[1]
		\Input Frozen classifier $f=h\circ g$, bias-audit set $\mathcal{D}$, patch size $s$, \#concepts $r$, step~size~$d$
		\Output Per-class ranked concept sets $\mathcal{R}$
		\State Initialize per-class ranked concept sets $\mathcal{R}\leftarrow\emptyset$
		\For{class $y\in\{1,\dots,C\}$}
			\State Collect $\mathcal{X}_y=\{x_i:\hat{y}(x_i)=y\}$ and patches $\mathcal{P}_y=\{\pi_s(x):x\in\mathcal{X}_y\}$
			\State Compute $A_y=g(\mathcal{P}_y)$ and NMF concepts $W_y$ via \cref{eqn:craft}
			\State Compute $\mathrm{FN}_y$ and $\mathrm{FP}_y$ on $\mathcal{D}$
			\For{$k=1,\dots,r$}
			\ForAll{$x\in\mathrm{FN}_y\cup\mathrm{FP}_y$}
			\State Form $a=g(x)$ and $a'=a-d\,\nabla_a L(h(a),y_i)$
			\State Solve NNLS in \cref{eq:nnls} to obtain $u_{y,k}(a)$ and $u_{y,k}(a')$, then $I_{y,k}(a)=\mathbf{1}[u_{y,k}(a)>0]$
		\EndFor
		\State Compute $S_{y,k}$ via \cref{eqn:score}
		\EndFor
		\State Add class-$y$ ranking to $\mathcal{R}$: concepts $w_{y,k}$ (columns of $W_y$) ranked by $S_{y,k}$ with their top-activating patches
		\EndFor\\
		\Return $\mathcal{R}$
	\end{algorithmic}
\end{algorithm}

\section{Dataset Construction Details}
\label[appendix]{app:datasets}

\paragraph{CMNIST.} In the biased training set each digit label is assigned a ``bias color'' that matches the class label with probability $0.95$. The \emph{unbiased} variant makes all color--label combinations equally likely, while the \texttt{shifted-bias} variant uses a different color-to-label assignment. A key advantage of CMNIST is the ability to modulate the bias correlation coefficient and generate biased versions of any sample by modifying its color, allowing for controlled experiments with known ground-truth bias directions.

\paragraph{Waterbirds.} Overlays birds from the Caltech-UCSD Birds (CUB) dataset~\cite{wah_caltech-ucsd_nodate} on Places~\cite{zhou_places_2018} backgrounds. The water/land background is correlated with the label with probability $0.95$ in training and $0.5$ in testing.

\paragraph{CelebA.} CelebA~\cite{liu2015faceattributes} contains aligned celebrity face images with $40$ binary attribute annotations. In the spurious-correlation setting it is used as a binary blond vs.\ not-blond prediction task with gender as the spurious attribute~\cite{Wu2023CelebA}; a biased training split is formed by subsampling so that target and spurious attribute are correlated at $0.95$. Evaluation reports group-wise and worst-group accuracy over the four (blond/not-blond)$\times$(female/male) combinations. The test split keeps a strong imbalance (e.g., only $180$ blond men against $9767$ not-blond women), which makes overall accuracy an unreliable standalone metric.

\section{CMNIST Results}
\label[appendix]{app:cmnist}

This section presents the full CMNIST results that complement the Waterbirds and CelebA results in the main text. These results were moved to the appendix because the simple and artificial nature of CMNIST makes it better suited as a controlled toy setting than as a headline benchmark.

\begin{figure*}[tb]
	\centering
    {}\hfill
	\begin{subfigure}[t]{0.5\textwidth}
		\centering
		\includegraphics[width=\linewidth]{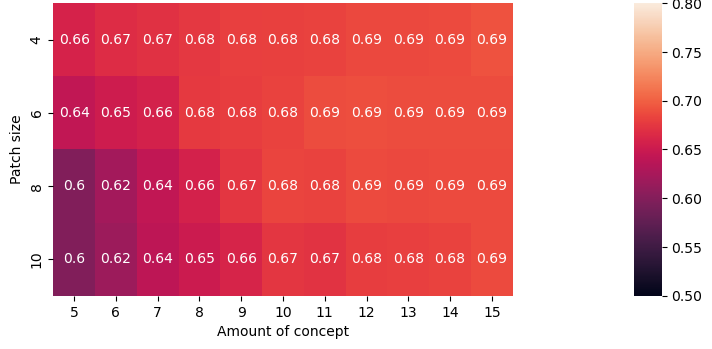}
		\caption{Biased model, biased audit}
		\label{fig:cmnist_align_a}
	\end{subfigure}\hfill
	\begin{subfigure}[t]{0.5\textwidth}
		\centering
		\includegraphics[width=\linewidth]{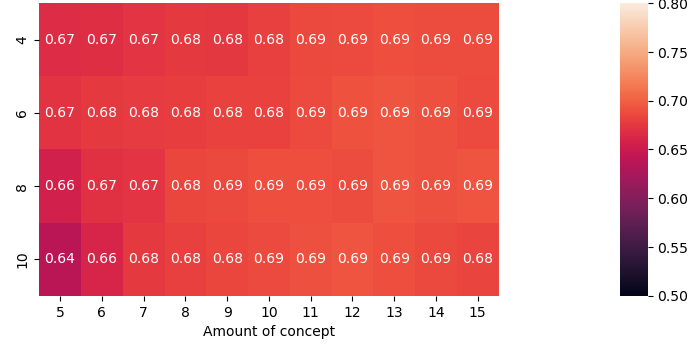}
		\caption{Biased model, unbiased audit}
		\label{fig:cmnist_align_b}
	\end{subfigure}\hfill{}\\
    {}\hfill
	\begin{subfigure}[t]{0.5\textwidth}
		\centering
		\includegraphics[width=\linewidth]{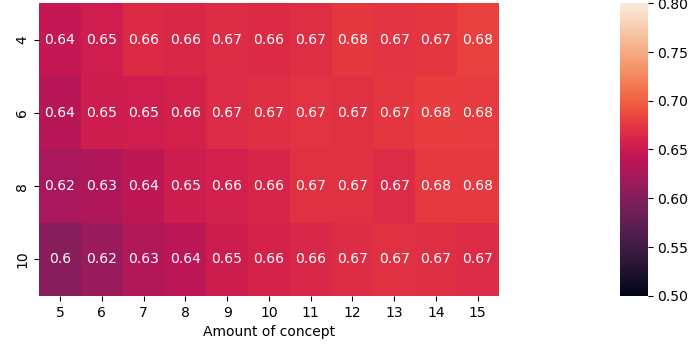}
		\caption{Biased model, differently biased audit}
		\label{fig:cmnist_align_c}
	\end{subfigure}\hfill
	\begin{subfigure}[t]{0.5\textwidth}
		\centering
		\includegraphics[width=\linewidth]{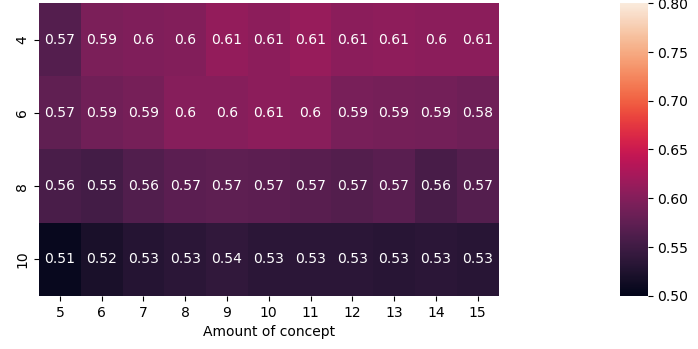}
		\caption{Unbiased model, biased audit}
		\label{fig:cmnist_align_d}
	\end{subfigure}\hfill{}\\
    {}\hfill
	\begin{subfigure}[t]{0.5\textwidth}
		\centering
		\includegraphics[width=\linewidth]{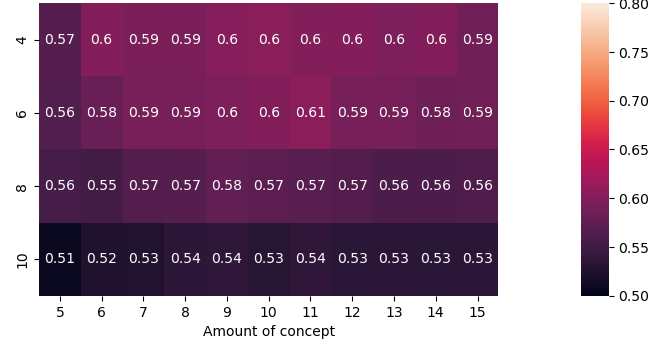}
		\caption{Unbiased model, unbiased audit}
		\label{fig:cmnist_align_e}
	\end{subfigure}
    \hfill{}
	\caption{CMNIST: cosine similarity between the estimated color-bias direction and the most aligned concept vector as a function of patch size and number of concepts. Values are computed over 10 runs on all 10 classes.}
	\label{fig:cmnist_alignment}
\end{figure*}

\subsection{Concept--Bias Alignment}

We sweep the patch size $s$ and number of concepts $r$. On CMNIST (\cref{fig:cmnist_align_a,fig:cmnist_align_b,fig:cmnist_align_c}), biased models on average always yield a strongly aligned concept across hyperparameters (minimum cosine similarity $0.6$ with low variance below $0.03$), and moderate settings ($r{\geq}9$, $s{\leq}8$) work best. While too few concepts hurt the emergence of a bias-aligned concept, too many concepts do not substantially improve the decomposition, plausibly because NMF then distributes bias information across multiple vectors rather than concentrating it in a single direction. Smaller patch sizes are generally preferred, likely because they focus patches on local evidence. The nature of the audit dataset has little influence on alignment overall, with effects mainly visible at low numbers of concepts. This suggests that non-negative concept decomposition can be used across different audit-set distributions. We note, however, that samples containing the bias attribute must still be present, even if the attribute is not correlated with the target label; otherwise, the decomposition cannot build patches that expose the bias attribute.

When the concept bank is built from an unbiased dataset (\cref{fig:cmnist_align_d,fig:cmnist_align_e}), concepts remain weakly aligned with the bias attribute. This confirms that the unbiased model does not rely on bias, as no strong bias direction emerges in its representation layer. It also suggests that non-negative concept decomposition is not prone to hallucinating spurious concepts when none are present.

\paragraph{Gradient step size.}
Sweeping $d$ on this setup, we find that the bias score becomes more reliable as the step size grows and plateaus without degrading at larger values; we therefore fix $d = 2\times10^{4}$ for all remaining experiments.

\subsection{Bias Score Validation}

\begin{figure*}[tb]
	\centering
    {}\hfill
	\begin{subfigure}[t]{0.3\textwidth}
		\centering
		\includegraphics[width=\linewidth]{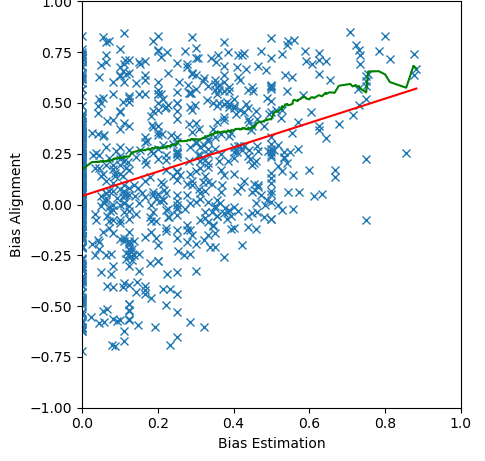}
		\caption{Biased model, biased audit}
		\label{fig:cmnist_score_a}
	\end{subfigure}\hfill
	\begin{subfigure}[t]{0.3\textwidth}
		\centering
		\includegraphics[width=\linewidth]{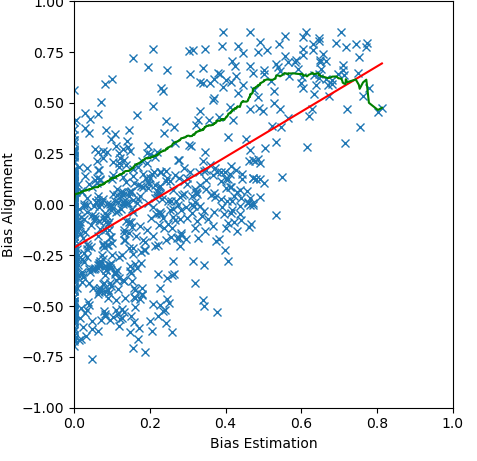}
		\caption{Biased model, unbiased audit}
		\label{fig:cmnist_score_b}
	\end{subfigure}\hfill
	\begin{subfigure}[t]{0.3\textwidth}
		\centering
		\includegraphics[width=\linewidth]{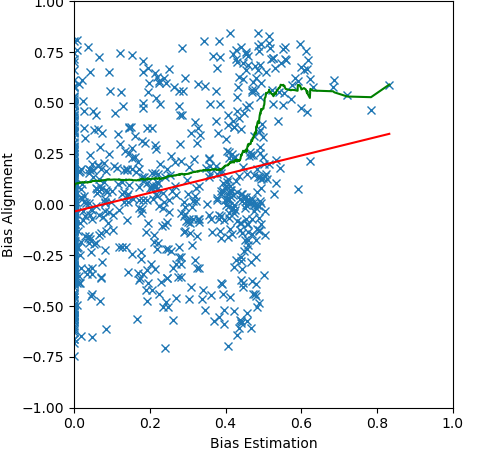}
		\caption{Biased model, differently biased audit}
		\label{fig:cmnist_score_c}
	\end{subfigure}\\
    {}\hfill
	\begin{subfigure}[t]{0.3\textwidth}
		\centering
		\includegraphics[width=\linewidth]{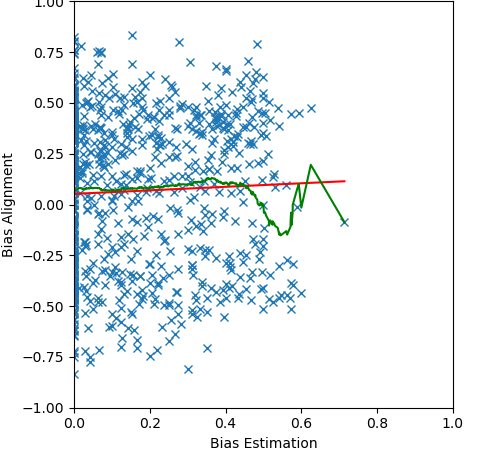}
		\caption{Unbiased model, biased audit}
		\label{fig:cmnist_score_d}
	\end{subfigure}\hfill
	\begin{subfigure}[t]{0.3\textwidth}
		\centering
		\includegraphics[width=\linewidth]{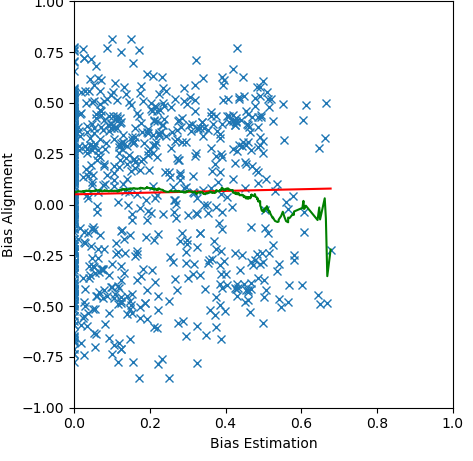}
		\caption{Unbiased model, unbiased audit}
		\label{fig:cmnist_score_e}
	\end{subfigure}\hfill{}
	\caption{CMNIST: bias score against bias-alignment score. Blue crosses represent concepts, red curves are linear regressions, and green curves show the average bias-alignment score of concepts above the current bias score threshold.}
	\label{fig:cmnist_bias_score_correlation}
\end{figure*}

On CMNIST, we examine the relation between bias score and bias alignment (\cref{fig:cmnist_bias_score_correlation}). On biased models (\cref{fig:cmnist_score_a,fig:cmnist_score_b,fig:cmnist_score_c}), alignment is widely spread at low bias scores, but this spread narrows as the score increases. Around a bias score of $0.7$, concepts consistently exceed $0.4$ in alignment, and the average alignment curve (green) confirms that concepts above this threshold have mean alignment above our $0.55$ criterion. This pattern holds across audit sets, although with weaker confidence when the audit set shares the same bias as the model, since informative error examples (false negatives without the spurious cue and false positives with the spurious cue) become rarer.

On unbiased models (\cref{fig:cmnist_score_d,fig:cmnist_score_e}), there is no strong correlation between the bias score and bias alignment, as expected: the model has not learned a spurious bias, so errors stem from random noise rather than systematic spurious attributes. No concept exceeds a bias score of $0.7$, reducing the risk of false bias identification. This holds even when the audit dataset itself is biased, indicating that the method is resilient to distributional properties of the audit set.

\subsection{Correlation with Ground-Truth Bias}

\Cref{tab:cmnist_correlation} reports the correlation between identified bias concepts and ground-truth bias labels for CMNIST. On biased models, identified concepts show significantly higher MCC with the color bias than other concept--bias pairs ($p < 0.001$). On unbiased models, no such pattern holds, confirming that the method does not hallucinate bias where none was learned.

\begin{table}[tb]
\centering
\small
\caption{CMNIST: correlation between identified bias concepts and ground-truth bias labels (cf.\ \cref{tab:correlation_metrics} for Waterbirds and CelebA).\label{tab:cmnist_correlation}}
\setlength{\tabcolsep}{.25em}
\renewcommand{\arraystretch}{1.05}
\resizebox{\linewidth}{!}{%
\begin{tabular}{@{}llS[table-format=1.1(1)]S[table-format=1.3]S[table-format=1.3]S[table-format=-1.3(3)]S[table-format=-1.3(3)]c@{}}
\toprule
{Train dataset} & {Audit bias} & {\#bias} & {$f_{\text{bias}}$} & {$f_{\text{other}}$} & {$\bar{\Phi}_{\text{bias}}$} & {$\bar{\Phi}_{\text{other}}$} & {$p$} \\
\midrule
\multirow{4}{*}{Biased CMNIST}   & biased       & 4.9 +- 1.8 & 0.959 & 0.919 &  0.418 +- 0.184 &  0.261 +- 0.204 & $0.001$ \\
& unbiased     & 5.7 +- 1.0 & 1.000 & 0.880 &  0.399 +- 0.154 & 0.156 +- 0.114 & $0.001$ \\
& shifted bias & 2.7 +- 1.1 & 0.963 & 0.856 &  0.479 +- 0.260 &  0.140 +- 0.136 & $0.001$ \\
\midrule
\multirow{2}{*}{Unbiased CMNIST} & biased       & 1.4 +- 1.1 & 0.027 & 0.004 & 0.027 +- 0.004 &  0.046 +- 0.027 & $0.857$ \\
& unbiased     & 2.1 +- 1.7 & 0.381 & 0.305 &  0.038 +- 0.017 &  0.043 +- 0.025 & $0.724$ \\
\bottomrule
\end{tabular}}
\end{table}

\subsection{Inference-Time Bias Mitigation}

\Cref{tab:cmnist_debias} reports the effect of suppressing identified bias concepts on CMNIST. On biased CMNIST, suppression does not consistently improve metrics and in some configurations decreases worst-group accuracy, particularly with the unbiased audit set, which is unexpected given this configuration's strong performance in the bias score validation. The ablation rows confirm this is not an artefact of removing concept capacity in general: random suppression degrades metrics further than targeted suppression in all conditions, so the identified directions remain preferentially bias-aligned even when their removal does not translate into worst-group gains. A plausible explanation is that on CMNIST the color attribute is almost linearly separable and tightly entangled with digit shape at our chosen rank, so removing the color-aligned direction also removes class-discriminative signal; this is consistent with the lower best-case worst-group accuracy observed relative to Waterbirds. Investigating multi-scale or recursive decompositions to disentangle these directions is left to future work. On unbiased CMNIST, suppressing identified concepts expectedly decreases all metrics, as the model did not rely on a spurious direction to begin with.

\begin{table}[tb]
\centering
\small
\caption{CMNIST: effect of suppressing identified bias concepts at inference time (cf.\ \cref{tab:debias_metrics} for Waterbirds and CelebA). ``--'' under \emph{Audit bias} denotes the base frozen model without mitigation.\label{tab:cmnist_debias}}
\setlength{\tabcolsep}{.25em}
\renewcommand{\arraystretch}{1.05}
\resizebox{\linewidth}{!}{%
\begin{tabular}{@{}llS[table-format=2.1(2.1)]S[table-format=2.1(2.1)]S[table-format=2.1(2.1)]@{}}
\toprule
{Train dataset} & {Audit bias} & {Accuracy} & {Worst-class acc} & {Worst-group acc} \\
\midrule
\multirow{7}{*}{Biased CMNIST}   & \texttt{--}  & 76.7 +- 0.5 & 64.6 +-  1.8 & 31.2 +-  5.1 \\
& biased       & 76.4 +- 2.7 & 61.9 +- 11.2 & 23.0 +- 12.5 \\
& biased ablation & 72.2 +- 3.1 & 54.6 +- 11.3 & 11.1 +- 9.5 \\
& unbiased     & 74.1 +- 2.4 & 52.0 +-  5.5 & 15.0 +- 12.8 \\
& unbiased ablation & 67.8 +- 4.9 & 36.5 +- 12.5 & 4.6 +- 7.0 \\
& shifted bias & 76.9 +- 0.7 & 64.2 +-  2.4 & 30.4 +-  7.1 \\
& shifted ablation & 74.1 +- 2.2 & 55.3 +- 8.8 & 19.1 +- 10.4 \\
\midrule
\multirow{5}{*}{Unbiased CMNIST} & \texttt{--}  & 95.8 +- 0.1 & 93.3 +-  0.5 & 88.3 +-  1.3 \\
& biased       & 94.8 +- 1.6 & 88.8 +-  7.1 & 82.4 +-  8.1 \\
& biased ablation & 95.0 +- 0.8 & 87.7 +- 6.7 & 79.3 +- 10.4 \\
& unbiased     & 94.0 +- 2.7 & 81.6 +- 20.0 & 75.5 +- 20.2 \\
& unbiased ablation & 95.0 +- 0.7 & 89.4 +- 5.1 & 81.5 +- 5.8 \\
\bottomrule
\end{tabular}}
\end{table}

\section{Comparison with Supervised Baselines}
\label[appendix]{app:baselines}

\Cref{tab:context} contextualizes our inference-time mitigation with supervised methods that retrain the model using group annotations. All reference numbers are from~\cite{kirichenko_last_2023} (ResNet-50); our Waterbirds model is a ResNet-18, our CelebA model is a ResNet-50.

\begin{table}[t]
\centering
\small
\caption{Worst-group accuracy (\%) of debiasing methods under different supervision and retraining requirements.  ERM = empirical risk minimization (standard training). $^\dagger$ResNet-50 numbers from~\cite{kirichenko_last_2023}; our Waterbirds model uses a ResNet-18.\label{tab:context}}
\setlength{\tabcolsep}{.3em}
\renewcommand{\arraystretch}{1.05}
\begin{tabular}{@{}lllcc@{}}
\toprule
Method & Bias labels & Retraining & Waterbirds & CelebA \\
\midrule
ERM (ours)                                                  & none      & none        & $45.9{\scriptstyle\pm1.5}$ & $43.4{\scriptstyle\pm2.5}$ \\
\textbf{Ours}                                               & \textbf{none} & \textbf{none} & $\mathbf{63.8{\scriptstyle\pm4.0}}$ & $\mathbf{53.8{\scriptstyle\pm4.1}}$ \\
\midrule
JTT$^\dagger$~\cite{liu_just_2021}                          & val       & full        & $86.7$  & $81.1$ \\
Group DRO$^\dagger$~\cite{sagawa_distributionally_2019}      & train+val & full        & $91.4$  & $88.9$ \\
DFR$^\dagger$~\cite{kirichenko_last_2023}                   & val       & last layer  & $92.9{\scriptstyle\pm0.2}$  & $88.3{\scriptstyle\pm1.1}$ \\
\bottomrule
\end{tabular}
\end{table}

Our method is the only one that requires neither bias labels nor parameter updates, operating entirely at inference time on a frozen model. The gap to supervised baselines reflects this strict constraint; methods that retrain with group-balanced data achieve higher absolute worst-group accuracy but presuppose label collection and model access that may not be available for deployed systems. Our approach provides a complementary, zero-cost diagnostic and mitigation handle that can be applied before deciding whether more invasive retraining is warranted.

\section{Bias Concept Interpretation}
\label[appendix]{app:interpret}

This appendix presents the full patch galleries and automatically generated labels for all bias concepts identified by our method on CMNIST, Waterbirds and CelebA. For each identified concept, we retrieve its top-100 most-activating patches from the bias-audit set and generate the 10 labels most associated with these patches (weighted by activation rank) using NOVIC~\cite{allgeuer2025unconstrained}, an open-vocabulary image classifier. Each subfigure shows a representative subset of the top-activating patches alongside the generated labels. Common concepts appear consistently across random seeds, while uncommon concepts emerge in only a fraction of seeds. Representative examples from both categories are included to illustrate the range of identified bias directions.

\begin{figure}[ht]
    \centering
    \begin{subfigure}{\textwidth}
        \centering
        \includegraphics[width=\linewidth]{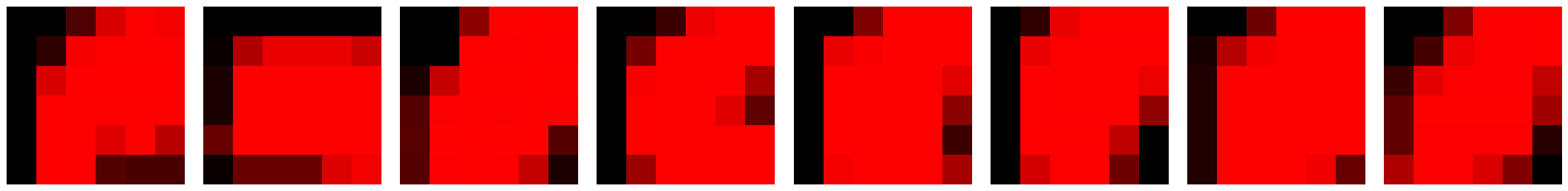}
        \caption{Bias for \textit{0}, ground-truth \textit{red}: \texttt{red hot}, \texttt{blood}, \texttt{red gram}, \texttt{red header}, \texttt{fdp}}
    \end{subfigure}
    \begin{subfigure}{\textwidth}
        \centering
        \includegraphics[width=\linewidth]{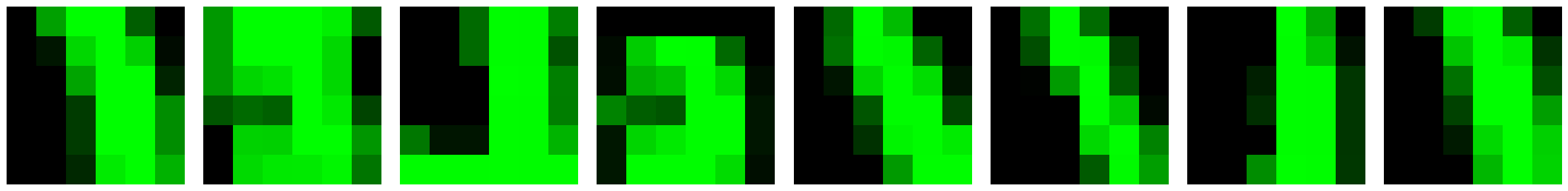}
        \caption{Bias for \textit{1}, ground-truth \textit{green}: \texttt{limelight}, \texttt{green}, \texttt{aurora}, \texttt{fume}, \texttt{rayon}}
    \end{subfigure}
    \begin{subfigure}{\textwidth}
        \centering
        \includegraphics[width=\linewidth]{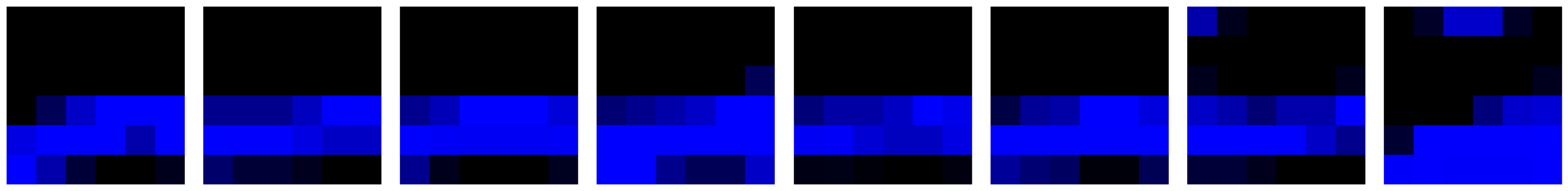}
        \caption{Bias for \textit{2}, ground-truth \textit{blue}: \texttt{blue point}, \texttt{blue}, \texttt{bleu}, \texttt{wavelet}, \texttt{bluepoint}}
    \end{subfigure}
    \caption{Top-activating patches of bias concepts for CMNIST with corresponding top-5 generated labels. Computed using biased model and similarly biased audit.}
    \label{fig:interpret_cmnist_app}
\end{figure}

\paragraph{CMNIST.}
CMNIST bias concepts (\cref{fig:interpret_cmnist_app}) align closely with ground-truth color bias. Regardless of the audit nature, as long as the model is biased, we observe that the ground-truth color bias label of each class appears systematically in the top-5 most appearing labels for bias concept of that class. One notable exception is class nine with the bias \textit{grey} for which no bias concept was ever recorded in any of the 10 experiments nor audit type. This seems to show that the \textit{grey} number is treated as a default by the model and therefore not strongly encoded. Interestingly, when the audit had unaligned bias, label three biased with \textit{yellow} in the model and \textit{red} in the audit also never displayed any bias concept.

\begin{figure}[ht]
    \centering
    \begin{subfigure}{\textwidth}
        \centering
        \includegraphics[width=.9\linewidth]{figures/concept_bamboo.png}
        \caption{Bias for \textit{landbird} (common): \texttt{bamboo}, \texttt{fishpole bamboo}, \texttt{phyllostachys}, \texttt{web}, \texttt{common bamboo}, \texttt{lesser bullrush}, \texttt{genus griselinia}, \texttt{damselfly}, \texttt{cue}, \texttt{yellow berry}}
    \end{subfigure}
    \begin{subfigure}{\textwidth}
        \centering
        \includegraphics[width=.9\linewidth]{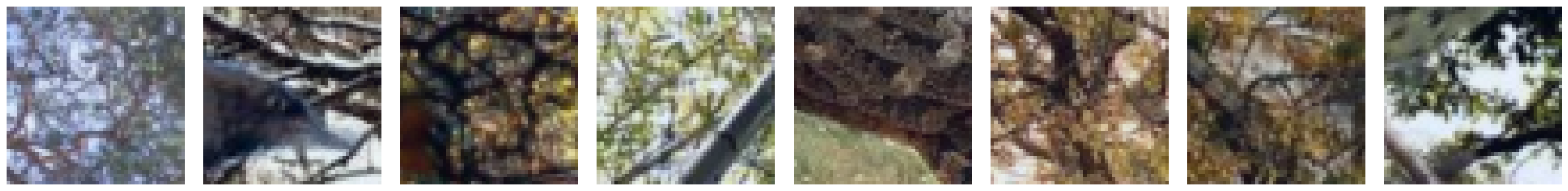}
        \caption{Bias for \textit{landbird} (uncommon): \texttt{fall}, \texttt{rain tree}, \texttt{birds eye maple}, \texttt{web}, \texttt{silverbush}, \texttt{black oak}, \texttt{blossom}, \texttt{brown oak}, \texttt{canopy}, \texttt{snow}. Appears in about half of the seeds; the fifth patch is a bird wing whose camouflage fits the ``leaf'' pattern.}
    \end{subfigure}
    \begin{subfigure}{\textwidth}
        \centering
        \includegraphics[width=.9\linewidth]{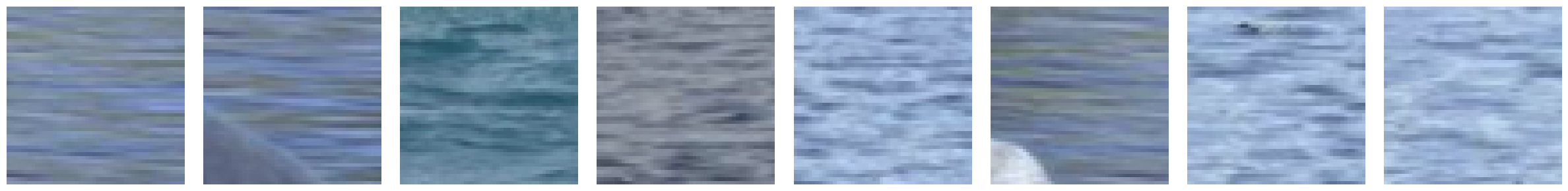}
        \caption{Bias for \textit{seabird} (common): \texttt{wavelet}, \texttt{wave}, \texttt{gray whale}, \texttt{blue}, \texttt{blue whale}, \texttt{sea}, \texttt{rippling}, \texttt{harbor seal}, \texttt{oceanic abyss}, \texttt{pilot whale}}
    \end{subfigure}
    \begin{subfigure}{\textwidth}
        \centering
        \includegraphics[width=.9\linewidth]{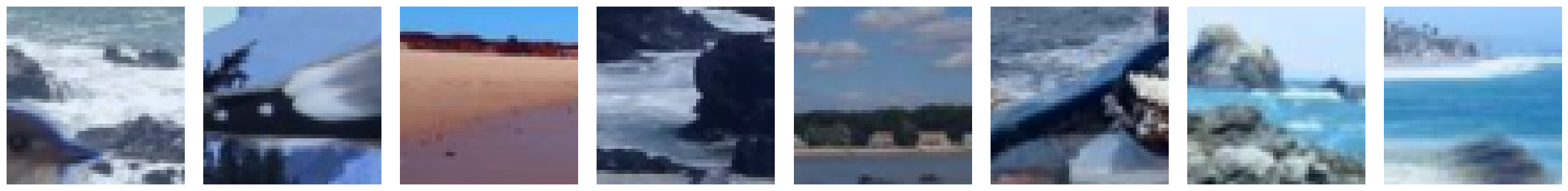}
        \caption{Bias for \textit{seabird} (uncommon): \texttt{coast}, \texttt{sea coast}, \texttt{wave}, \texttt{hill}, \texttt{mountain}, \texttt{beach}, \texttt{coast boykinia}, \texttt{gray whale}, \texttt{breakwater}, \texttt{tidal wave}. Appears in about 1/5th of the seeds; the bird--sea boundary is often confused for a coast.}
    \end{subfigure}
    \caption{Top-activating patches of common and uncommon bias concepts for Waterbirds with corresponding top-10 generated labels.}
    \label{fig:interpret_water_app}
\end{figure}

\paragraph{Waterbirds.}
The Waterbirds bias concepts (\cref{fig:interpret_water_app}) align closely with the ground-truth background bias. The two common concepts capture the dominant spurious cues: dense vegetation for \textit{landbird} (a) and open water for \textit{seabird} (c). The uncommon concepts reveal subtler patterns: autumnal foliage (b) and coastal boundaries (d). The foliage concept appears in about half of the seeds and occasionally includes a bird wing whose camouflage pattern resembles a leaf texture. The coastal concept appears in roughly one-fifth of seeds and often confuses the bird--sea boundary for a coastline. These results confirm that the NMF decomposition reliably separates background-related directions from bird-related ones.

\begin{figure}[ht]
    \centering
    \begin{subfigure}{\textwidth}
        \centering
        \includegraphics[width=.9\linewidth]{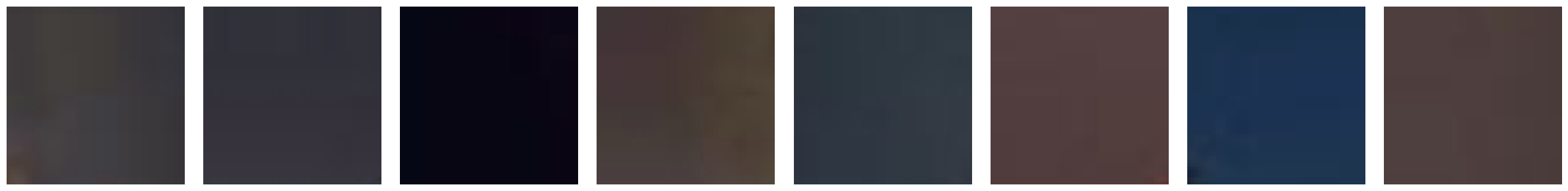}
        \caption{Bias for \textit{not blond}: \texttt{black}, \texttt{sky}, \texttt{maroon}, \texttt{blue sky}, \texttt{gray}, \texttt{blue}, \texttt{indigo}, \texttt{brown ash}, \texttt{black ash}, \texttt{dark chocolate}. This uncommon concept shows the model decision is influenced by background color; it correlates fairly well with the \texttt{men} bias (MCC $0.318$) and matches societal stereotypes.}
    \end{subfigure}
    \begin{subfigure}{\textwidth}
        \centering
        \includegraphics[width=.9\linewidth]{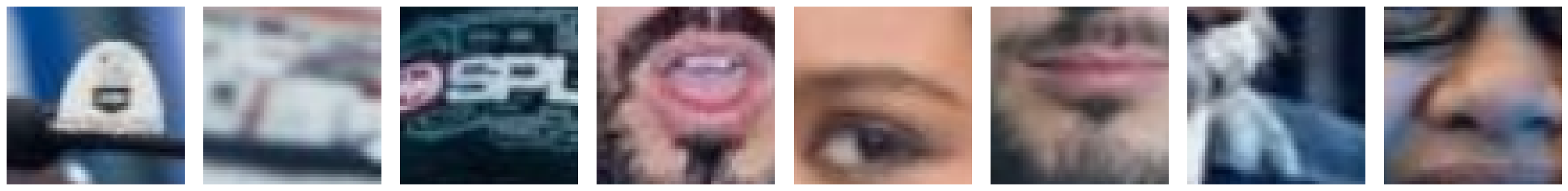}
        \caption{Bias for \textit{not blond}: \texttt{lip}, \texttt{eyebrow}, \texttt{lip rouge}, \texttt{lip gloss}, \texttt{mouth}, \texttt{eyelash}, \texttt{brow}, \texttt{facial hair}, \texttt{sideburn}, \texttt{nose}. Appears with variation fairly commonly and illustrates the distributed nature of the gender concept.}
    \end{subfigure}
    \begin{subfigure}{\textwidth}
        \centering
        \includegraphics[width=.9\linewidth]{figures/concept_blond_hair.png}
        \caption{Bias for \textit{blond}: \texttt{human hair}, \texttt{hair}, \texttt{body hair}, \texttt{hair space}, \texttt{hair wig}, \texttt{brown ash}, \texttt{false hair}, \texttt{marble wood}, \texttt{fiber}, \texttt{golden thread}. A common concept featuring blond hair; the presence of long hair suggests the direction may capture shape rather than color, illustrating that color and shape information co-locate on the same patches.}
    \end{subfigure}
    \begin{subfigure}{\textwidth}
        \centering
        \includegraphics[width=.9\linewidth]{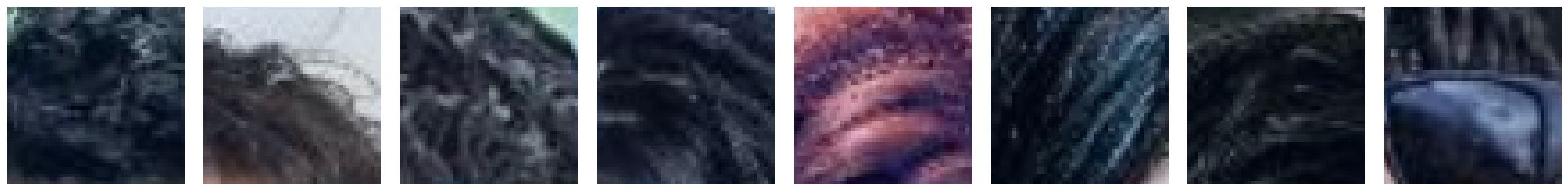}
        \caption{Bias for \textit{not blond}: \texttt{hair}, \texttt{human hair}, \texttt{hair space}, \texttt{black ash}, \texttt{fiber}, \texttt{black}, \texttt{hair wig}, \texttt{nebula}, \texttt{raven}, \texttt{fur}. Features only dark hair; the bias-mitigation experiments indicate this concept does encode bias although the exact aspect is ambiguous.}
    \end{subfigure}
    \begin{subfigure}{\textwidth}
        \centering
        \includegraphics[width=.9\linewidth]{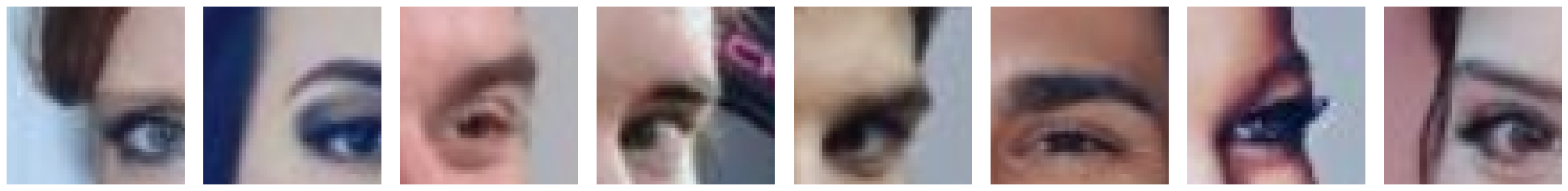}
        \caption{Bias for \textit{not blond}: \texttt{eyebrow}, \texttt{eyelash}, \texttt{brow}, \texttt{make up}, \texttt{lash}, \texttt{human eye}, \texttt{eye}, \texttt{sideburn}, \texttt{eyeshadow}, \texttt{human nose}. Concentrates on the eyes; plausibly bias-aligned since eye make-up is correlated with gender in CelebA.}
    \end{subfigure}
    \begin{subfigure}{\textwidth}
        \centering
        \includegraphics[width=.9\linewidth]{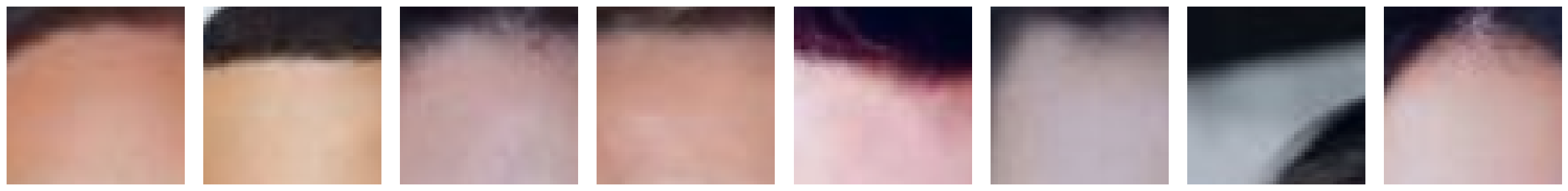}
        \caption{Bias for \textit{not blond}: \texttt{hair}, \texttt{forehead}, \texttt{sideburn}, \texttt{eyebrow}, \texttt{brow}, \texttt{human hair}, \texttt{hair space}, \texttt{body hair}, \texttt{eyelash}, \texttt{hair wig}. Features foreheads; derived from patches classified as \textit{not blond} it contains mostly dark hair and captures hairline/forehead cues that can vary by gender.}
    \end{subfigure}
    \caption{Top-activating patches of common and uncommon bias concepts for CelebA with corresponding top-10 generated labels.}
    \label{fig:interpret_celeb_app}
\end{figure}

\paragraph{CelebA.}
The CelebA bias concepts (\cref{fig:interpret_celeb_app}) are more diverse and harder to interpret, reflecting the distributed nature of the gender cue in this dataset. Unlike Waterbirds, where the bias is spatially localized in the background, gender-related information on CelebA is spread across multiple facial regions. The identified concepts span background color and darkness (a), lower-face features such as lips and facial hair (b), hair texture and color for \texttt{blond} (c) and \texttt{not blond} (d), eye-region features including make-up cues (e), and forehead and hairline structure (f). While no single concept correlates strongly with the binary gender attribute, collectively these directions capture decision-relevant spurious information: suppressing them improves worst-group accuracy by 10.4~percentage points (\cref{tab:debias_metrics}). This pattern is consistent with the observation in \cref{sec:correlation} that gender on CelebA cannot be represented as a single concept direction under a patch-based decomposition.

\clearpage
\section{Extended Related Work}
\label[appendix]{app:related}

Reliance on spurious correlations is a common failure mode in vision, where predictors perform well on the training distribution but fail to generalize when the underlying correlations differ at deployment~\cite{geirhos_shortcut_2020}. Surveys show such shortcuts persist across model families and datasets, and are addressed with methods that vary in supervision and access to the training pipeline~\cite{ye_clever_2025}. Prior work broadly either (i) mitigates bias given known spurious or group information, (ii) improves robustness without explicit bias supervision during training, or (iii) discovers spurious cues post-hoc when the bias is unknown.

\subsubsection{Bias-aware Mitigation.}
When spurious attributes or groups are available, approaches reduce reliance via regularization~\cite{bahng_learning_2020}, adversarial training~\cite{kim_learning_2019}, or representation-space erasure~\cite{belrose_leace_2023-1}. Other methods use concept supervision to probe where a bias is encoded and how it affects predictions~\cite{correa_efficient_2024,kim_interpretability_2018}, or intervene along a specified artifact direction using augmentation, projection, or gradient penalties~\cite{anders_finding_2022,dreyer_hope_2024}. While effective with correct bias signals, these methods assume the spurious attribute is known and typically require training-time access; in contrast, we operate post-hoc on a frozen model without bias labels. MaskTune reduces shortcut reliance via post-training fine-tuning on saliency-masked inputs~\cite{asgari_masktune_2022}, and subspace-projection methods such as JSE remove specified spurious concepts~\cite{holstege_removing_2024}, but both still rely on optimization and/or spurious supervision to define what to suppress.

\subsubsection{Bias Discovery Without Bias Labels.}
A related line of work discovers latent groups or failure slices without subgroup labels, aiming to explain performance gaps and support robust learning. Loss-based schemes upweight examples misclassified by an auxiliary model~\cite{nam_learning_2020,liu_just_2021}, and environment-discovery or subgroup-identification methods recover pseudo-groups from model behavior or representations~\cite{pezeshki_discovering_2024,zhang_discover_2024,zare_frustratingly_2024}; slice-discovery methods generate pseudo-labels for downstream mitigation~\cite{ghosh_ladder_2025,olesen_slicing_2025}. Group-free evaluation and model selection are also non-trivial~\cite{labonte_group_2024}, motivating proxy validation and pseudo-labeled validation sets~\cite{tsirigotis_group_2023,ghaznavi_exploiting_2025}. These approaches localize failures, but generally do not yield a direct, class-conditional representation-space direction that can be ranked and targeted.

\subsubsection{Automatic Concept Discovery and Post-hoc Debiasing.}
Concept discovery methods extract human-interpretable concepts from activations without concept annotations~\cite{lee_concept-based_2024,ghorbani_towards_2019,fel_craft_2023}, and have been used to improve robustness by reweighting or rebalancing discovered concepts during training~\cite{arefin_unsupervised_2024}. Related work studies factorization-based concept ranking~\cite{kowal_understanding_2024-1}, hierarchical or patch-level concept bottlenecks~\cite{panousis_coarse--fine_2024}, and the faithfulness and reliability of concept directions for intervention~\cite{bhusal_face_2025,pahde_navigating_2024-1}. Other post-hoc pipelines cluster explanation maps or use vision-language models to propose candidate biases~\cite{anders_finding_2022,chakraborty_exmap_2024,kim_discovering_2024,paduraru_conceptdrift_2024}, but often require manual validation and do not provide a concrete class-conditional direction that can be directly suppressed.

\subsubsection{Post-hoc Mitigation with Frozen Models.}
Several methods reduce spurious reliance after ERM training without group labels, including last-layer retraining on a class-balanced split (DFR)~\cite{kirichenko_last_2023}, sub-network extraction~\cite{le_out_2024}, activation-level erasure~\cite{he_eva_2024}, and data pruning~\cite{mulchandani_severing_2024}. Vision-language work similarly studies inference-time debiasing under deployment constraints \cite{gerych_bendvlm_2024,hirota_saner_2024}. While aligned in motivation, these approaches still perform optimization, pruning, or model edits, and typically do not expose an interpretable direction; we instead identify class-conditional spurious concept directions in a frozen model to support downstream intervention without updating parameters.

\end{document}